\documentclass[conference]{IEEEtran}
\IEEEoverridecommandlockouts

\usepackage{graphicx}
\usepackage{epstopdf}

\usepackage{cite}
\usepackage{amsmath,amssymb,amsfonts}
\usepackage{algorithmic}
\usepackage{textcomp}
\usepackage{xcolor}
\usepackage{subfig}
\usepackage{layouts}
\usepackage{siunitx}

\usepackage[shortlabels]{enumitem}
\usepackage[ruled,vlined]{algorithm2e}

\usepackage{hyperref}
\hypersetup{colorlinks,allcolors=black}

\def\BibTeX{{\rm B\kern-.05em{\sc i\kern-.025em b}\kern-.08em
		T\kern-.1667em\lower.7ex\hbox{E}\kern-.125emX}}

\newcommand{\mN}{\mathbb{N}}

\newcommand{\vx}{\textbf{x}}  
\newcommand{\vz}{\textbf{z}} 

\makeatletter
\def\ps@IEEEtitlepagestyle{%
  \def\@oddfoot{\mycopyrightnotice}%
  \def\@evenfoot{}%
}
\def\mycopyrightnotice{%
  {\footnotesize978-1-7281-8942-0/20/\$31.00 \textcopyright 2020 IEEE\hfill}
  \gdef\mycopyrightnotice{}
}

\begin{document}
	
	\title{Offline Auto Labeling: BAAS
	\thanks{A part of this work is funded by German Federal Ministry of Education and Research (BMBF) and Federal Ministry for Economic Affairs and Energy (BMWi) through Radar4FAD project with the project grant no. 16ES0560.}
	}
	
	\author{\IEEEauthorblockN{ Stefan Haag$^*$, Bharanidhar Duraisamy$^*$, Felix Govaers$^\#$, Wolfgang Koch$^\#$, Martin Fritzsche$^*$ J\"urgen Dickmann$^*$}
		\IEEEauthorblockA{ (*) Research and Developpement, Mercedes-Benz AG, Germany, Email: firstname.lastname@daimler.com and\\
			(\#) Dept. for Sensor Data and Information Fusion, Fraunhofer FKIE, Germany, Email:firstname.lastname@fkie.fraunhofer.de \\}
	}
	
	\maketitle
	
	\begin{abstract}
		This paper introduces BAAS, a new Extended Object Tracking (EOT) and fusion-based label annotation framework for radar detections in autonomous driving. Our framework utilizes Bayesian-based tracking, smoothing and eventually fusion methods to provide veritable and precise object trajectories along with shape estimation to provide annotation labels on the detection level under various supervision levels. Simultaneously, the framework provides evaluation of tracking performance and label annotation. If manually labeled data is available, each processing module can be analyzed independently or combined with other modules to enable closed-loop continuous improvements. The framework performance is evaluated in a challenging urban real-world scenario in terms of tracking performance and the label annotation errors. We demonstrate the functionality of the proposed approach for varying dynamic objects and class types.   
	\end{abstract}
	
	\section{Introduction}
	
	Automated L3 and Autonomous driving L4 (AD) systems require robust environment perception algorithms to handle complex traffic scenarios in urban, semi-urban, and highways of various countries. Under this premise, artificial Intelligence (AI) - deep learning approaches either with supervised or semi-supervised learning capability play a vital role: at the moment, especially in object detection applications, but in the future, as well, in higher dimension decision making and inference activities. Earlier years in the automotive domain, deep learning methods were applied primarily on camera data and tailored for object detection applications. However, in recent years, deep learning methods are investigated for tracking and fusion applications on data obtained through various active and passive sensors \cite{Krebs2018, Wang2019}.
	\begin{figure}[htb]
		\centering
		\includegraphics[width=\columnwidth]{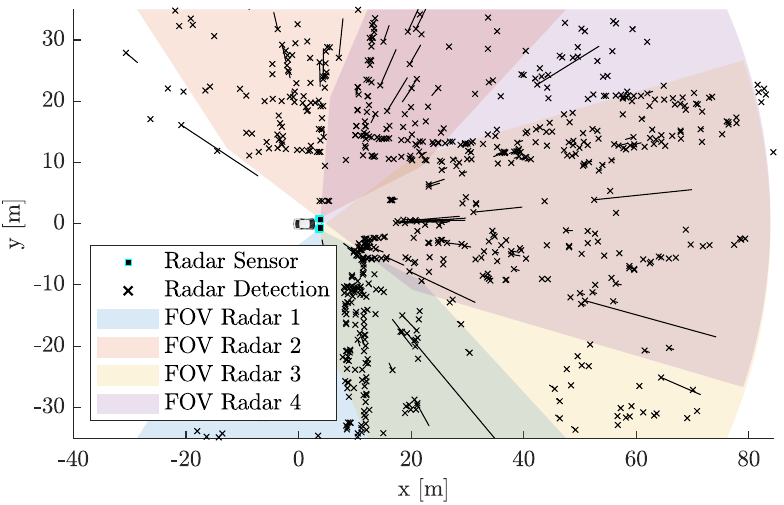}
		\caption{Measurement data from Sensor Platform (SP), radar detections with range-rate measurement, colorized backgrounds indicate the sensors' overlapping fields of view}
		\label{fig:FOV}
	\end{figure}	
	Thankfully, many open source data sets for the automotive domain are available to test AI algorithms for their performance \cite{Yin2017} and to statistically evaluate new approaches. However, the vast majority of the published data sets are recorded mostly under the right weather conditions, and available open-source data under challenging weather conditions are still scarce. A quick survey on the open-source data sets shows that principally they consist of stereo and mono camera, lidar, and GPS data. Nevertheless, actual radar detections are not published in any of these well known open-source data sets \cite{Kang2019}.
	\begin{figure*}[htb]
		\centering
		\includegraphics[trim=0 220 0 0,width=\textwidth]{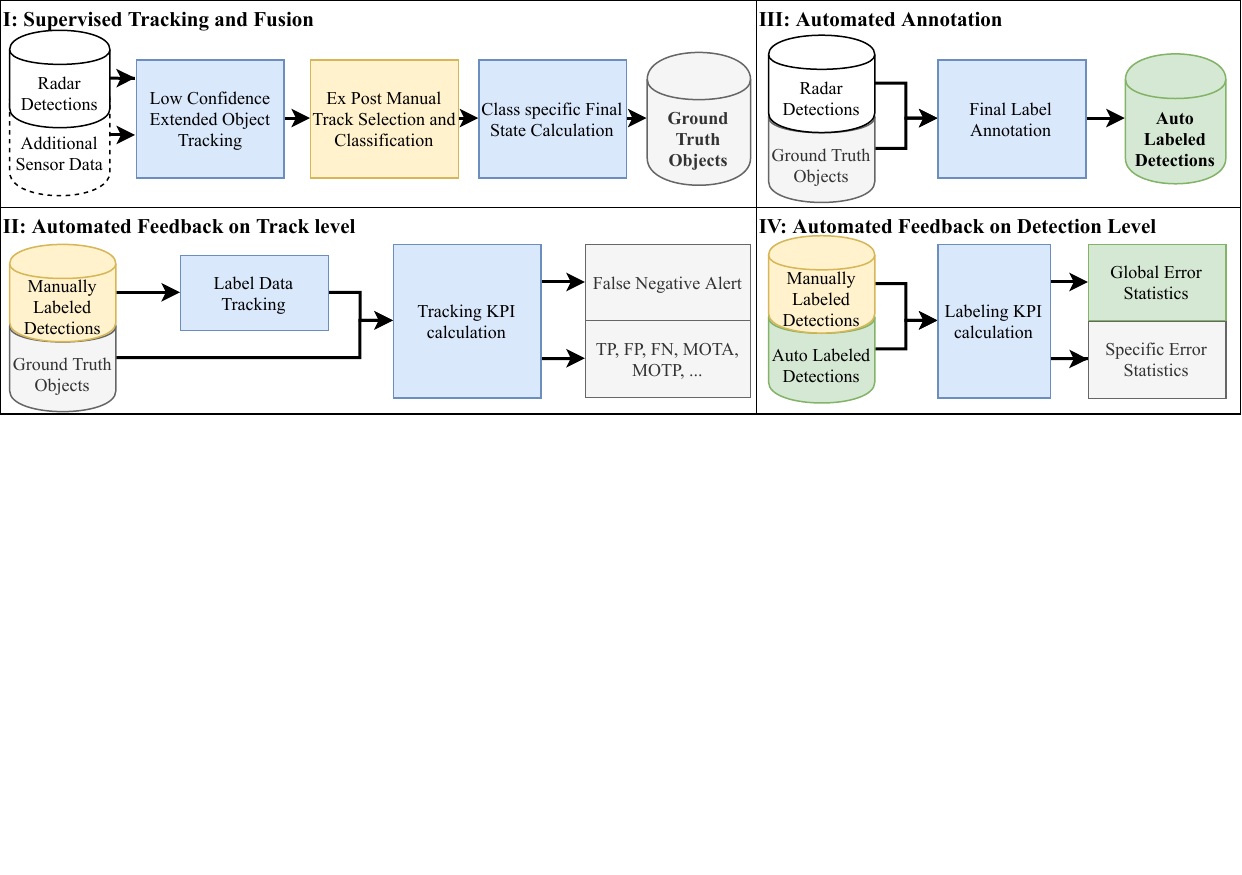}
		\caption{BAAS: Bayesian Assisted Automated Annotation System's open-loop process modules are depicted. Database symbols represent data sets. Automated processes are marked in blue, manual in yellow tiles. Internal results are marked in gray, final results in green. Module I and II provide results on the object level, module III, and IV on the detection level. BAAS module I operates with sensor data. Input data is integrated with sensor models to obtain the annotation hypothesis with our tracking and fusion system. Annotation hypothesis history is internally referred and saved. False positives and track degeneracy must be supervised through a human labeler to obtain verified tracks. Given the corrections and class information from the manual labeling, the ground truth is approximated for each occurring object. BAAS module II evaluates results from module I against manually labeled data to ensure continuous optimization of module I. BAAS module III calculates label annotations based on given object ground truth. The corresponding evaluation is provided from module IV}
		\label{fig:Process}
	\end{figure*}	
    However, automotive radars are an essential active sensor for autonomous driving applications because of a fundamental property that radar operates on a spectrum that is known to be more stable under challenging conditions such as darkness, fog, or rain. Furthermore, radar provides direct velocity measurements based on the Doppler principle that enables fast object detection for all kind of maneuver planning, collision evasion, and emergency braking systems. In our opinion, there could be two different reasons that lead to a lack of public available radar data. First, sharing radar data allows inferences about the sensors supplier-specific signal processing capability, which many companies would not be willing to share. Second, annotating radar data is a complicated and time-consuming task that requires manual labeling from human labelers with radar expertise. It is costly and scales with the size of the data set.
	
	In order to apply deep learning approaches in radar-based environment perception, massive labeled radar data sets are required. The lack of labeled data led us to exploit our Bayesian tracking and fusion system \cite{Haag2019a} to reduce the effort and aid in the data annotation task. This paper presents a novel approach to utilize the Bayesian tracking and fusion system that calculates and matches the annotation hypothesis from low-level radar detections primarily to objects that are in a state of motion, also known as dynamic objects. We named our approach as Bayesian Assisted Annotation System (BAAS), and it would be referred under the acronym throughout this paper. BAA approach could be applied to an already annotated data set to obtain the time tracked data series of the dynamic objects to obtain the temporal, spatial, and other correlations among the traffic participants. We have derived an Unscented Interacting Multiple Model (UIMM) smoother to improve the accuracy and robustness of the position, kinematic, and extension state information. BAAS provides several annotation hypotheses in an iterative process and is designed with caution not to provide full automation yet, as safety standards have to be ensured in the learning process and to reduce false positives as an outcome of the learning process. However, manual human annotation task complexity is reduced to supervision and correction mode for eventual object management tasks. It is possible to extend BAA to integrate the feedback from the correction task to adopt necessary filters. Despite the technical feasibility and viability of BAAS, we have decided to keep the system as an open-loop system due to complying with our organization's AI technical compliance.    
	
	The paper is organized as follows, section \ref{sec:sota} discusses related work. Section \ref{sec:pm} introduces our proposed labeling automation system: BAAS, containing multiple EOT and smoothing, post trajectory verification, and the final label annotation. In section \ref{sec:er}, the experimental results based on low-level radar data with ground truth evaluation are presented. Finally, section \ref{sec:Conclusion} concludes the paper with our summary.
	
	\section{Related Work}
	\label{sec:sota}
	
	Acquiring ground truth information and information about provided measurements from real-world road users is challenging \cite{Scheiner2019}. Ex post determinations are, in most cases, impossible. Nevertheless, there is a growing demand for real-world data, where each measurement, respectively detection, is annotated with a label referring to the object providing the measurement. Existing approaches for sensor data labeling can be divided into three different approaches.
	
	The first possible approach to label measurements is to draw minimal bounding boxes around each visible object manually. If very accurate are available, the bounding boxes denote information about an object's position, motion state, size, and orientation.     Labeling with bounding boxes can be simplified by labeling only specific frames and interpolating the boxes into the enclosed frames. Precise bounding boxes cannot be determined from automotive radar measurements due to the reduced angular resolution. Therefore, the object extent can not be determined with manual labeling for automotive radar measurements.
	
	The second possible approach is automated labeling with devices for ground truth gathering, e.g., GNSS devices, as applied in \cite{Scheiner2019} for single vulnerable road users on high-resolution radar data, or in \cite{Haag2018} for a single car on automotive radar data or in \cite{Behrendt2019} on lidar data. 
	However, the application is limited, as each object has to be equipped with an individual device. In practice, the number of objects is limited because the scenarios have to be arranged, and the GNSS devices are costly.
	
    The possible third approach is the usage of partially automated systems that provide labeling proposals near the actual results so that only a review from the labeler with small corrections is required, similar to the approach chosen in this paper.  Again, camera data was first considered. In addition to non-automotive applications \cite{Bianco2015, Hanyu2018}, the first was made in labeling pedestrians in movie data \cite{Nino2016, Tumas2018} with object detectors. Tacking-based approaches for label annotation on camera data were presented in \cite{Comaschi2014}. In \cite{Wang2018}, a tracking-based labeling approach is introduced, with a focus on the impact of manual corrections on accuracy.
	
	\section{Proposed Method}
	\label{sec:pm} 	
	
	Unlike human labelers, BAAS can not provide annotation labels for radar measurements without object information. For that reason, object ground truth information is calculated first in a separate module. The annotation is calculated in an independent module so that differently gathered ground truth information can be included in the labeling framework. For each module, an evaluation and feedback module is required that ensures continuous optimization of each automated process step. An overview of BAAS modules, their processes, and their dependencies is shown in Fig. \ref{fig:Process}. At least, radar measurements have to be provided to utilize BAAS. Additional information such as measurements from additional sensors or AI gained results can be integrated to improve the performance. In this paper, only radar measurements are utilized since this is the most challenging case.   
	
	\subsection{Problem Definition}
	
    BAAS is conceived for post-processing recorded measurement from real-world traffic scenarios, where only the dynamic objects are instantiated. Measurements from static objects and clutter are considered as one set. 
    The dynamic object labeling problem is divided into two subtasks.
    First, all dynamic objects visible to the radar sensors have to be determined. Implicitly, the cardinality $m_k$ of objects per scan time $t_k$ and the number of objects $M \in \mN$ is determined. 
    For each object, $\tau_k^l, \ l = 1,..., M$, the centroid position, kinematic information is tracked along with the object extent for its lifespan, $(t_{K_1^l}, t_{K_2^l})$.
    Second, annotation weights $\rho_k^{j,l} \in [0,1]$ indicating if a measurement can be assigned to a track which or not are calculated, for each detection of a sensor scan $Z_k = \{ \vz_k^j \}_{j = 1}^k$. Binary labeling, as performed by manual labeling, allows only $\rho_k^{j,l} \in \{0,1\}$ and clear assignments of the detection to its object.
	
	\subsection{Extended Object Tracking}
	
	EOT for BAAS is subjected to different requirements than regular EOT. 
	\begin{figure}[bht]
		\includegraphics[width=\columnwidth]{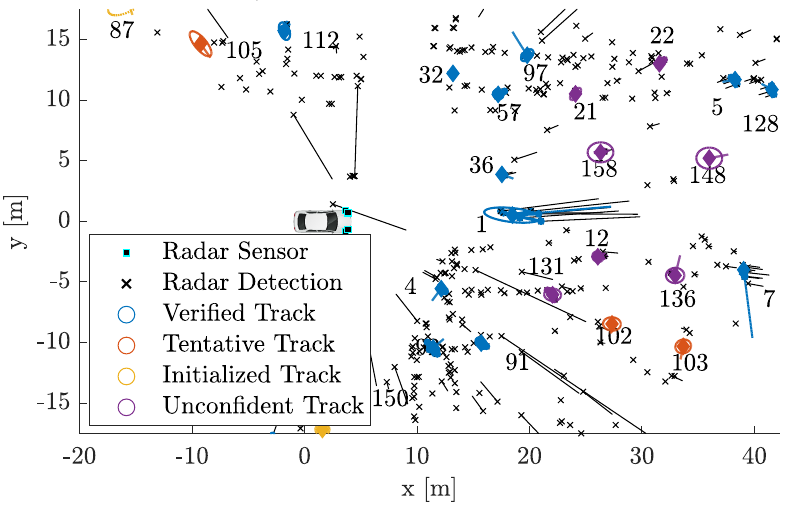}
		\caption{Tracking result showing all existing track hypotheses and their ID. A car (1) and 4 of 5 pedestrians are tracked as (4, 5, 36, 128) verified tracks. A Cyclist is tracked with initialized track (147) and the last pedestrian with unconfident track (12). The remaining tracks are false positives}
		\label{fig:TrackingEx}
	\end{figure}
	False positives tracks can be suppressed by manual supervision, but false negative tracks require expensive manual labeling. A real-time application is not required. Therefore, EOT is performed with the maximal number of tracking hypotheses. The hypotheses are propagated with Adaptive Clustering (AC) \cite{Haag2019a} that solves both, data association and clustering. AC is again combined with the UIMM Random Matrix Model (RMM) filter \cite{Haag2019a}, but with clustering parameters providing smaller clusters as a track respectively cluster merge can be corrected in the manual supervision step. The BAAS specified extended object tracking result of the urban scenario, previously shown in Fig \ref{fig:FOV}, is shown in Fig. \ref{fig:TrackingEx}. A large number of false positive tracks are required to ensure coverage of all objects. Low confident tracks are established to track objects that provide (mostly) single detections so that the coverage of all objects, especially pedestrians, is more likely. 
	
	\subsection{Manual Supervision}
	
	The supervisor's work in BAAS is to extract the correct tracks from the created tracking hypotheses after EOT was performed on all time steps.
	\begin{figure}[bht]
		\centering		
		\includegraphics[width=\columnwidth]{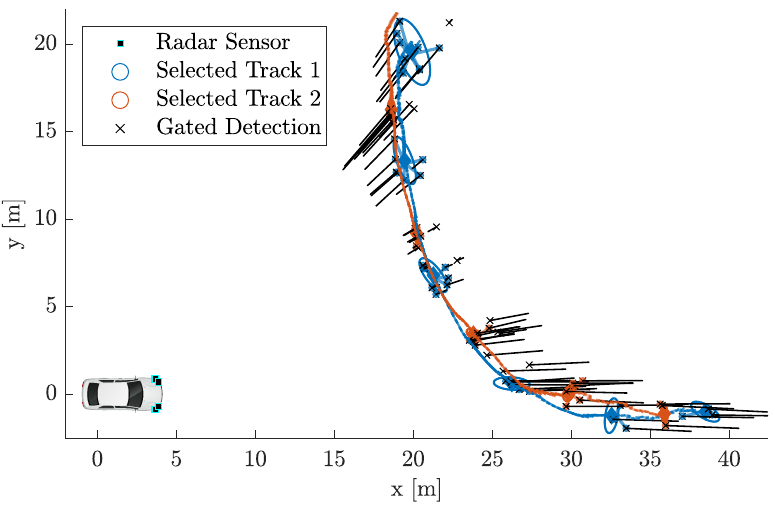}
		\caption{Obtained trajectories of two tracks selected to cover a single car. The car enters the SP's FOV from the left and turns left. Pedestrians cross the street between the observed car and the SP. Object extents and gated detections are shown for every 20th time step. Associated measurements are indicated with colored lines to the object's concurrent centroid}
		\label{fig:TrackingPath}
	\end{figure}
	The supervisor can not interfere in the tracking process. Thereby, the computational effort has no impact on the duration of the supervision effort. Instead of manually selecting detections, the supervisor has to observe EOT results over time and select the true positive tracks by its id. Furthermore, he has to merge split tracks by denoting the track ids of divided tracks. Due to AC and partial occlusions, track splits are propagated over several time steps. Observation over time and comparison with the documentary camera images give the supervisor evidence when tracks have to be merged.	Further options of manual track manipulation are possible, but they have to be operating on the track level and they should be controllable on a graphical user interface to keep the task simple. 
	
	\subsection{Final State Calculation}
	
	The trajectories of the veritable tracks $\tau^{(l_1,...,l_m)}$ are obtained by combining all measurement assigned to the tracks $l_1,.,l_m$. 
	In the case of merged tracks, the measurements of several tracks are combined. No direct track to track fusion is performed, the track states are re-calculated. Thus, complex cases such as multiple simultaneous existing tracks or temporary track losses are covered. The obtained trajectories are smoothed and extension estimators have to be re-calculated for the smoothed trajectories. Further post-processing is handled differently depending on the class type of the tracked object. Bicycles, cars and trucks are assumed to be rigid bodies oriented in the direction of movement. Average values are calculated for length and width. The orientation is adjusted if the object is moving. The correct orientation of the ellipse is essential to establish a highly reliable data association in dynamic scenarios. However, the RMM can not distinguish between size and orientation because of the shape representation as a matrix. Thus turning maneuvers can lead to shape deformations. For that reason, better association results are obtained if the ellipse is aligned with the object's moving direction for cyclists, cars and trucks if the object moves fast enough
	\begin{align}
	\alpha_k^l = \arctan \left( \frac{\dot{x}_k^l}{{y}_k^l} \right) \text{ , if } \vert \vert \left( \dot{x}_k, \dot{y}_k \right) \vert \vert > \eta_v
	\end{align} 
	Furthermore, object length and width are assumed to be static. They are calculated as the weighted average
	\begin{align}
	\left\{l^l,w^l\right\} = \sum_{k = K_0^l}^{K_1^l}  \frac{\nu_k^l}{ \sum_{i = K_0^l}^{K_1^l} \nu_i^l } \left\{l_k^l,w_k^l\right\}
	\end{align}
	\begin{figure}[bht]
		\centering		
		\includegraphics[width=\columnwidth]{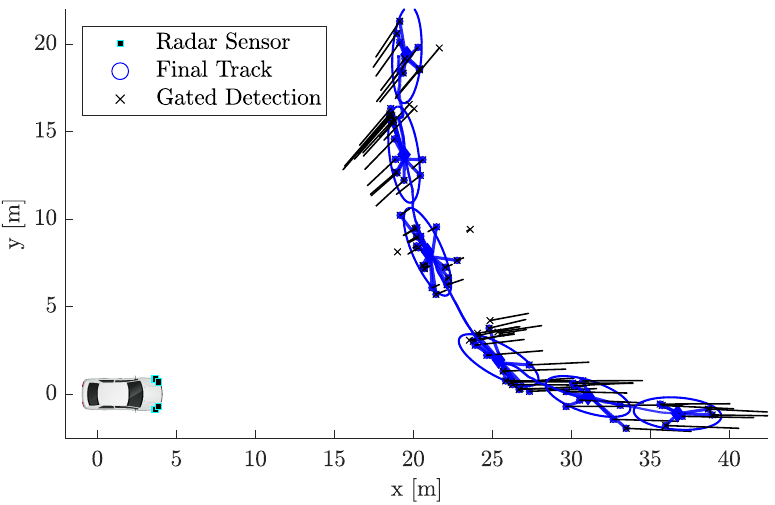}
		\caption{Final trajectory and the fixed object extension with aligned orientation for every 20th time step. Annotations are indicated by blue lines}
		\label{fig:TrackingPathFinal}
	\end{figure} 
	Pedestrians are assumed to have fixed sizes, but the orientation is independent of the moving direction. For them, only average length and width are calculated. The extent of Pedestrian Groups and Other objects are not corrected as they may vary over time. The labeler can set minimum and maximum sizes depending on each class type. Thus, significant size estimation errors caused by partial occlusion or high distance to the sensor are prevented. The result of the previous example in Fig. \ref{fig:TrackingPath} is shown in Fig. \ref{fig:TrackingPathFinal}.	
	
	\subsection{Automated Feedback on Track Level}
	
	Since object information is not provided from the manually labeled data sets, the object information is reconstructed by applying the same EOT filter on the manually clustered detections. The cluster to track association is provided from manual labeling. This process is called label data tracking in Fig. \ref{fig:Process}. Label data tracking results are utilized as benchmark. The module examines if objects are completely missed. If this is the case, the utilized base clustering parameters \cite{Haag2019a} have to be adjusted. Furthermore, the module provides tracking performance evaluation of the ground truth objects.	For this application, the multi object tracking accuracy (MOTA) \cite{Bernardin2008} is calculated, where a sum of tracking errors: the number of false positives (FP$_k$), false negatives (FN$_k$), mismatches (MM$_k$) is divided by the number of true positives (TP$_k$) and subtracted from 1   
	\begin{align}
	    \text{MOTA} = 1 - \frac{ \sum_{k= 0}^T \left( \text{FN}_k + \text{FP}_k + \text{MM}_k  \right)}{ \sum_{k= 0}^T \text{TP}_k }
	\end{align}
	MOTA$=1$ indicates the optimal object coverage. Decreasing MOTA values indicated decreasing tracking performance.
	The multiple object tracking precision (MOTP) determines the average distance between the ground truth objects and the label data tracking results.
	MOTP is denoted by the number of matched tracks from the two different tracking results M$_k$ and the centroid error for each matched couple $\tilde{\vx}_k^i$.
	Small MOTP values are required to ensure good performance in the dataset annotation.
	\begin{align}
	\text{MOTP} = \frac{ \sum_{k= 0}^T \sum_{i = 1}^{\text{M}_k} \tilde{\vx}_k^i }{ \sum_{k= 0}^T \text{M}_k }
	\end{align}  
	
	\subsection{Automated Annotation}
	
	\begin{figure}[bht]
		\centering
		\includegraphics[width=\columnwidth]{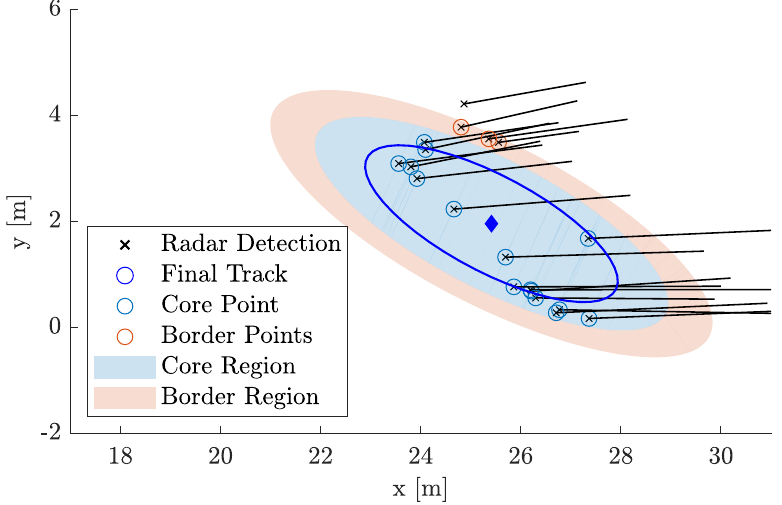}
		\caption{Automated annotation for a single scan. The object state is shown as dark blue ellipse. The acceptance regions, depending of the track position, track extent and measurement noise are shown as colored areas, core region in light blue and border region in red. Assigned detections are circled} 
		\label{fig:corePtsSel}
	\end{figure} 
	The final measurement annotation is calculated with the previously calculated ground truth objects. It is derived from AC core point selection \cite{Haag2019a}.
	A measurement $\vz_k^j$ is assigned to Track $\tau^l$ if the core point criteria \cite{Haag2019a} is fulfilled with its pseudo measurement $\hat{\vz}_k^l$ 
	\begin{align}
	    \left( \vz_k^j - \hat{\vz}_k^l \right) \left( z X_k^l + R_k^j \right)  \left( \vz_k^j - \hat{\vz}_k^l \right) \le \underbrace{\chi_3^2( \alpha )}_{\text{Core}} + \underbrace{\eta_B \left( \xi_{k}^l \right)}_{Border}
	\end{align}
	Core and border areas, along with associated measurements, are shown in Fig. \ref{fig:corePtsSel}. Finding a suitable function for the additional threshold $\eta_B \left( \xi_{k \vert k} \right)$ is challenging, but they can be learned with results from BAAS module IV.
	
	\subsection{Automated Feedback on Detection Level}
	
	The automated feedback module allows statistical optimization of the final association process. The previously calculated track to track association from module II is utilized to compare annotated detections with the manually labeled clusters. Module IV requires manually labeled data sets to calculate the numbers of true negative (TN), TP, FN and FP labeled detections. With this value precision and recall, values are calculated with which the arithmetic mean, the so-called F1 score is denoted. The border region function $\eta_B(\vx_k)$ is optimized regarding this values. Furthermore, the occurrence of label annotation errors can be further evaluated by examining the occurrence region or the class type of the tracked object to detect the weaknesses of the whole framework.
	
	\section{Experimental Results}
	\label{sec:er}
	
	\begin{figure}[bht]
		\centering
		\includegraphics[width=\columnwidth]{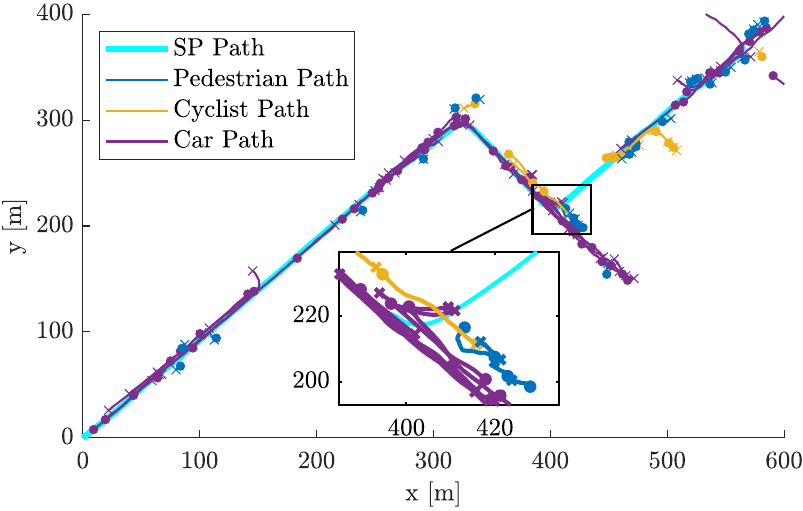}
		\caption{Object Paths in test scenario. Object birth positions are drawn with a dot-symbol and the last observed positions with an x. The SP path is shown in cyan with start at the origin} 
		\label{fig:pathPlot}
	\end{figure}
	BAAS evaluation is divided into two parts, similar to the proposed modules II and IV.
	BAAS is tested in an exemplary urban scenario which is shown in Fig. \ref{fig:pathPlot} with the sensor set up shown in Fig. \ref{fig:FOV}. The real-world scenario is 2 min long. It was recorded in regular urban traffic. Object paths, obtained from label data tracking, are shown in Fig. \ref{fig:pathPlot}. Fifty-nine cars, thirteen cyclists twenty-six pedestrians occur in the scenario. The performance is evaluated after each of the five consecutive BAAS processing steps. 
	\begin{enumerate}[1:]
		\item After BAAS specified low confident EOT
		\item After automated track selection
		\item After supervised track selection and merging
		\item After smoothing and shape re-calculation
		\item After size and position corrections 
	\end{enumerate}	. 
	
	\subsection{Evaluation on Track Level}
	
\begin{figure}[bht]
		\centering
		\includegraphics[width=\columnwidth]{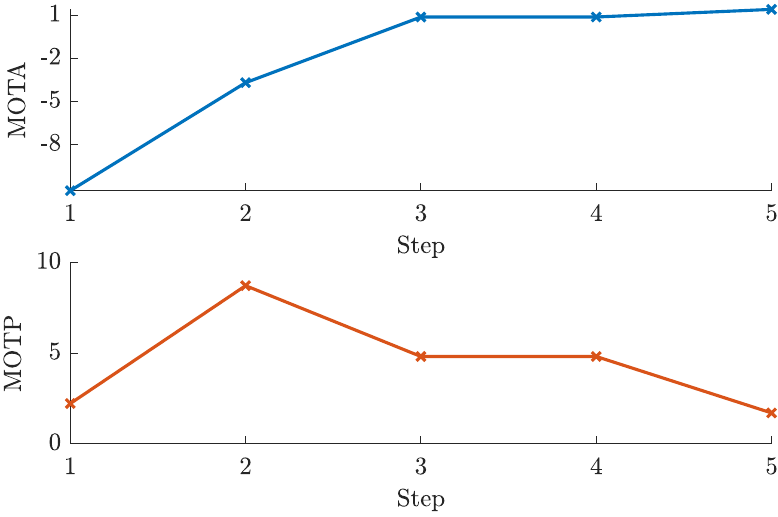}
		\caption{BAAS tracking performance in example scenario evaluated after each process step. The MOTP values are calculated with the euclidean distance with position and velocity values} 
		\label{fig:evalPlotT}
	\end{figure}

	Fig. \ref{fig:evalPlotT} shows MOTA and MOTP values of the example scenario after each step.
	MOTA values after the first step are shallow, due to the high amount of false positive tracks. However, the values increase in every step, especially with manual supervision, where all the false positives tracks are deleted. MOTA after step 5 is slightly below one, because track deletion times are not identical.Excellent performance values on the track level is a necessary precondition for automated label annotation.
	MOTP becomes only relevant if good association accuracy is achieved. The low MOTP value after step can be explained with the higher amount of tracks in general and the reduced track stability. Fig. \ref{fig:evalPlotT} shows that the last correction step is necessary to obtain significantly lower precision values. Thereby, correct label annotation becomes more likely.
	
	\subsection{Evaluation on Detection Level}
	
	For this paper the binary label case is evaluated as it is performed from the manual labelers as well. No distinction between core and border points is made. Miss-matched detections occur very rarely compared to falsely associated clutter measurements. Therefore, they are added to the false positives.
	\begin{figure}[bht]
		\centering
		\includegraphics[width=\columnwidth]{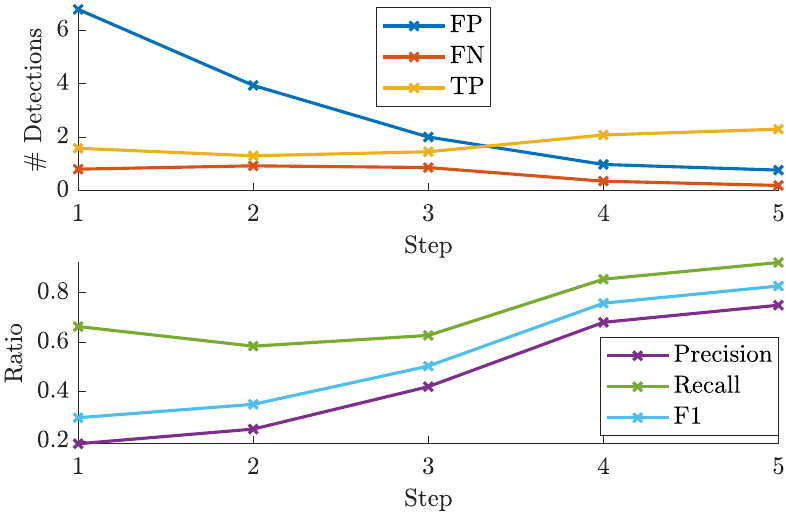}
		\caption{BAAS label annotation performance. TOP: mean FP, FN and TP numbers per scan. Bottom: mean precision, recall and F1 score after each BAAS processing step} 
		\label{fig:evalPlotD}
	\end{figure}	
	The numbers of true negative measurements is omitted. The mean number of true negative detections per scan is 120 after step 5.
	Fig. \ref{fig:evalPlotD} shows that false positives are more common than false negatives. 
	It shows that Step 4, shape re-calculation and ellipse alignment, increase the number of true positive association from $1.4$ to $2.1$. In the last step, the number of false positives per time step is increased by $0.2$. In the first two steps, FPs are tolerated, but still, the number is $0.5$ false positive labeled detection after the last processing step. The precision after step 5 is $81$\%. In every step, the number of false positives is significantly lower, but the difficulty lies in reducing the number of false positives detections and false negatives detections simultaneously. The average number of false negatives is $0.2$ per scan after step 5, and the automated labeling recall is $92$\%, leading to an average F1-score of $0.87$.
	
	\section{Conclusion}
	\label{sec:Conclusion}
	
	We have introduced BAAS, an EOT and fusion based label annotation framework. Model intelligence is utilized to automate the annotation task as far as possible and thereby simplify the effort of the annotation task on radar detections significantly. This framework has two inherent advantages: First, reliable object information about the size and kinematic state of each object is obtained. Second, a data-driven, statistical, EOT performance is obtained from the introduced framework. Through the analysis of the annotation error, we showed that the framework achieves sufficient accuracy in a challenging urban scenario. Analyzing the annotation errors allows further improvement in the tracking performance and, thereby, improvement of the performance of the labeling framework.	But most essentially, we showed that EOT integrated into a labeling framework is helping to simplify the label annotation task to radar detections.      

    \bibliographystyle{ieeetr}
	\bibliography{library2} 
	
\end{document}